%% file: main.tex
\documentclass[runningheads]{llncs}

\usepackage{eccv}

\usepackage{eccvabbrv}
\usepackage{graphicx}
\usepackage{booktabs}
\usepackage{multirow}
\usepackage{amsmath}
\usepackage{amssymb}
\usepackage{xcolor}
\usepackage{enumitem}
\usepackage{subcaption}
\usepackage{float}
\usepackage[accsupp]{axessibility}
\usepackage{tabularx}
\usepackage[skins,breakable]{tcolorbox}
\usepackage[table]{xcolor}

\usepackage[pagebackref,breaklinks,colorlinks,citecolor=eccvblue]{hyperref}

\usepackage{orcidlink}

\newcommand{\method}{\texttt{ArtiCAD}}

\begin{document}

\title{ArtiCAD: Articulated CAD Assembly Design\texorpdfstring{\\}{~}via Multi-Agent Code Generation}
\titlerunning{ArtiCAD}

\author{
Yuan Shui\inst{1} \and
Yandong Guan\inst{1} \and
Zhanwei Zhang\inst{2} \and
Juncheng Hu\inst{1} \and
Jing Zhang\inst{1} \and
Dong Xu\inst{3} \and
Qian Yu\inst{1}\textsuperscript{\dag}
}

\authorrunning{Y. Shui et al.}

\institute{
School of Software, Beihang University \and
Zhejiang University \and
The University of Hong Kong\\[4pt]
\textsuperscript{\dag} Corresponding author: \href{mailto:qianyu@buaa.edu.cn}{qianyu@buaa.edu.cn}\\
Project page: \url{https://shui-yuan.github.io/articad/}
}

\maketitle

\input{sec/abstract_V2}
\input{sec/introduction_V2}
\input{sec/related_work}
\input{sec/representation}
\input{sec/method}
\input{sec/experiments}
\input{sec/applications}
\input{sec/conclusion}

\bibliographystyle{splncs04}
\bibliography{main}

\end{document}

%% file: sec/abstract_V2.tex
% ===============================================================
% ABSTRACT
% ===============================================================
\begin{abstract}
Parametric Computer-Aided Design (CAD) of articulated assemblies is essential for product development, yet generating these multi-part, movable models from high-level descriptions remains unexplored. 
To address this, we propose \method{}, the first training-free multi-agent system capable of generating editable, articulated CAD assemblies directly from text or images. 
Our system divides this complex task among four specialized agents: Design, Generation, Assembly, and Review. One of our key insights is to predict assembly relationships during the initial design stage rather than the assembly stage. 
By utilizing a Connector that explicitly defines attachment points and joint parameters, \method{} determines these relationships before geometry generation, effectively bypassing the limited spatial reasoning capabilities of current LLMs and VLMs. 
To further ensure high-quality outputs, we introduce validation steps in the generation and assembly stages, accompanied by a cross-stage rollback mechanism that accurately isolates and corrects design- and code-level errors. Additionally, a self-evolving experience store accumulates design knowledge to continuously improve performance on future tasks. 
Extensive evaluations on three datasets (ArtiCAD-Bench, CADPrompt, and ACD) validate the effectiveness of our approach. We further demonstrate the applicability of \method{} in requirement-driven conceptual design, physical prototyping, and the generation of embodied AI training assets through URDF export.

% 具身引用

\keywords{CAD Generation \and Articulated Objects \and Multi-Agent }
\end{abstract}

%% file: sec/introduction_V2.tex
% ===============================================================
% INTRODUCTION
% ===============================================================
\section{Introduction}
\label{sec:intro}

\begin{figure}[t]
  \centering
  \includegraphics[width=\linewidth]{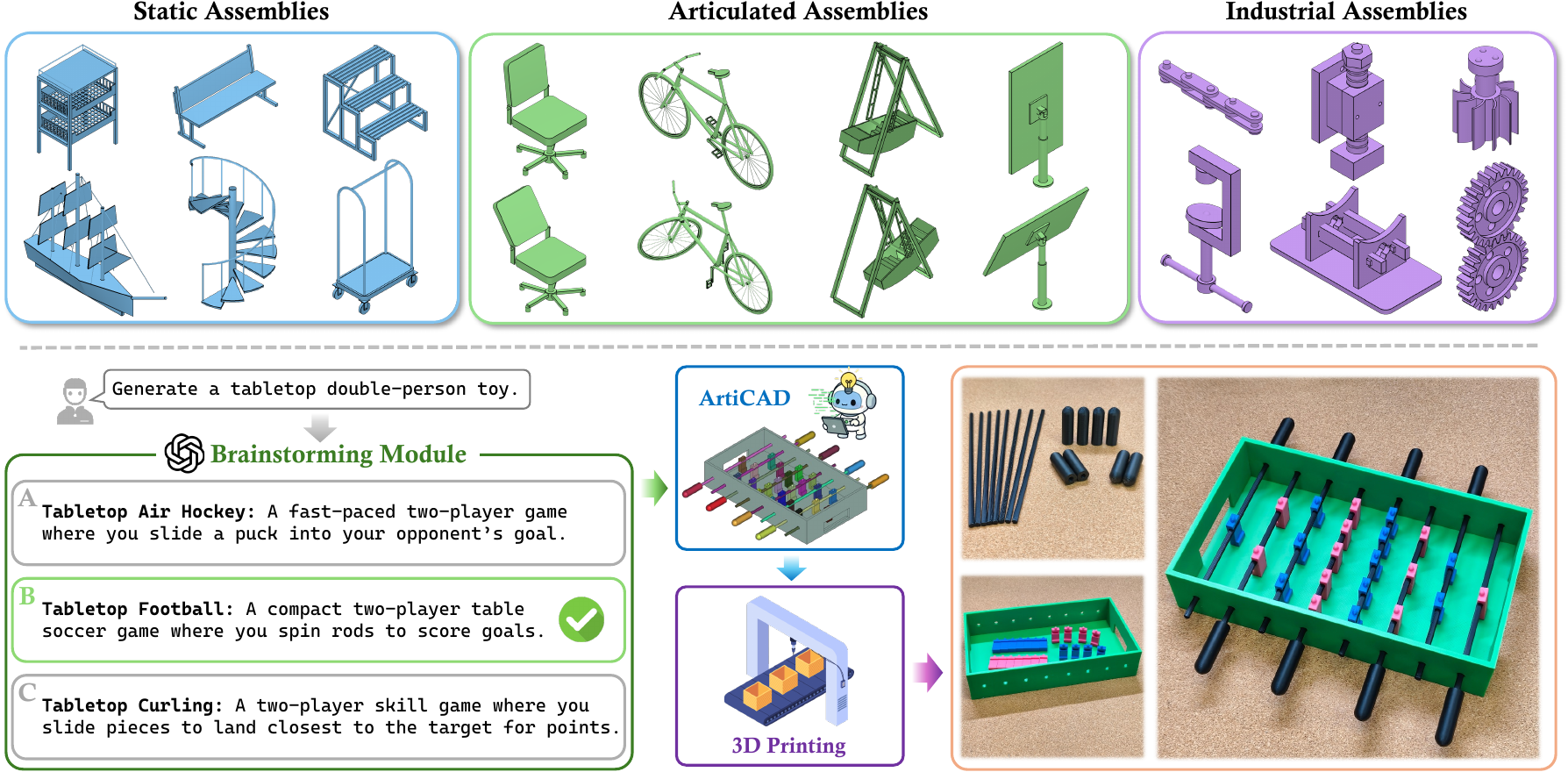}
  \vspace{-2em}
  \caption{\textbf{Top:} 
  CAD Assemblies generated by \method{} across three task categories: Static, Articulated, and Industrial. All outputs are editable. 
  \textbf{Bottom:} 
   An example application. Given a user requirement, \method{} generates an articulated CAD assembly with functional components (\eg, an enclosure, rods/handles, and player pieces). The components are then fabricated using a 3D printer (Bambu Lab P1S) and assembled into a working tabletop football prototype. (Photos taken by the authors.)}
%  Demonstration of an end-to-end pipeline from a text requirement to a functional physical artifact. Given the prompt ``Generate a tabletop double-person toy'', an external Brainstorming Module proposes several design schemes. After the user selects the ``Tabletop Football'' option, \method{} generates an articulated CAD assembly with meaningful part structures (e.g., enclosure, rods/handles, and player pieces). The components are then fabricated using a Bambu Lab P1S 3D printer and assembled into a functional tabletop football prototype.}}
  \label{fig:teaser}
  \vspace{-1.5em}
\end{figure}

Parametric Computer-Aided Design (CAD) underpins modern product development. Recent LLM-based methods generate single parametric parts from text or images with reasonable fidelity~\cite{wang2025cadgpt,xu2024cadmllm,guan2025cadcoder}. Most real-world products, however, are \emph{assemblies}: multiple components joined through typed joints with prescribed degrees of freedom (DOF). Generating articulated CAD assemblies from high-level descriptions remains a largely unsolved problem.

%As part count grows, output length and cross-part consistency demands increase accordingly.
%A coordinate or dimension error in one part may propagate until the final assembly is inspected, with no intermediate checkpoint to catch it.
%Current LLMs also struggle with 3D spatial reasoning---inferring how coordinate systems and dimensions align across independently generated components remains unreliable.
%These compounding difficulties make a monolithic approach impractical.
Existing work addresses only parts of this problem.
On one side, CAD code generation methods~\cite{khan2024text2cad,rukhovich2025cad,guan2025cadcoder,xie2025textcadquery,seekcad2025,alrashedy2025cadcodeverify,preintner2025evocad,govindarajan2026cadmium} produce single static parts; they have no mechanism for joints or multi-part hierarchies. 
On the other side, prior methods for articulated object reconstruction~\cite{liu2025singapo,le2025articulateanything,cao2026physxanything,liu2026pact,liu2025artgs,shen2026gaussianart} infer geometry and joint parameters from images or videos, typically producing non-editable mesh or 3D Gaussian representations. 
While these models successfully learn to reconstruct specific moving parts (\eg, a rotating hinge on a cabinet door), their ability to generalize is limited to the object categories seen during training.   %which predominantly consist of furniture.  
%For example, even though a bicycle's steering column and pedals also operate via standard revolute joints—sharing the exact same kinematic relationship as those doors and lids—existing data-driven models fail to generate them simply because the bicycle category itself is absent from their training distribution. 
%In contrast, ArtiCAD successfully generalizes to such unseen categories, generating diverse articulated structures alongside static and industrial assemblies (e.g., the bicycle shown in \cref{fig:teaser}). 
Furthermore, because these methods output fixed surface shapes rather than parametric CAD programs, their results lack the construction history necessary for downstream engineering and design iterations.

\begin{figure}[t]
  \centering
  \includegraphics[width=\linewidth]{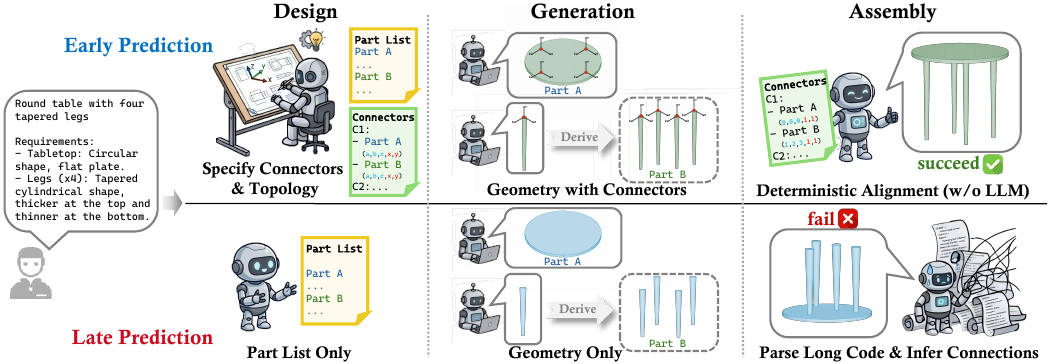}
  \vspace{-1.5em}
  \caption{Early vs.\ late assembly relationship prediction. \textbf{Top}: early prediction (ours) specifies connectors at design time; 
  %each part generation module receives an explicit geometric target and generates its part independently; 
  assembly reduces to deterministic frame alignment. \textbf{Bottom}: deferring connection decisions to assembly stage forces a second planning pass that must parse all generated code, infer coordinate systems, and resolve cross-part dimensions---a task with long context and high failure rate.}
  \label{fig:pipeline_comparison}
  \vspace{-1.5em}
\end{figure}

To bridge this gap, we propose \method{}, a training-free multi-agent system that enables articulated CAD assembly generation from text or images, as shown in Fig.~\ref{fig:teaser}(Top).
Creating articulated assemblies introduces two key challenges. 
\textbf{First}, the task involves coupled sub-problems: requirements analysis, structural design, part generation, and final assembly. 
Because a single LLM or VLM struggles to handle such complex tasks reliably, we use a multi-agent system to divide the work among specialized roles. 
\textbf{Second}, and most crucially,  predicting how components connect is a unique challenge for this task compared to single CAD part generation. 
Current LLMs and VLMs struggle with precise 3D spatial reasoning. 
If we wait until the assembly stage to predict relationships—forcing the model to look at independently generated parts and guess how their coordinate systems and surfaces align—the system faces a massive search space and often fails (as illustrated in Fig.~\ref{fig:pipeline_comparison}).
Therefore, a key insight of \method{} is to perform assembly relationship prediction in the \emph{design stage} rather than the \emph{assembly stage}. Similar to following a blueprint rather than blindly fitting puzzle pieces together, early prediction sets up a connection plan before any 3D shapes are generated. 
By explicitly defining these connections in advance, we separate the connection planning from the shape generation. This avoids the LLM's spatial reasoning limits during the final assembly.

Specifically, our multi-agent system has four agents. A \textbf{Design Agent} receives user input (text, image, or both) and outputs a structured plan containing part specifications and assembly relationships, which we call \emph{connectors}.
Each connector defines a named attachment point carrying a local coordinate frame, a semantic label, and a joint type with motion limits.
Next, \textbf{Generation Agents} produce each part independently as a FreeCAD Python script~\cite{freecad2024} based on the Design Agent's plan.
%\textcolor{red}{Symmetric or repeated parts need not be generated independently:  \emph{derive mechanism} applies a deterministic transform to the source geometry and its connectors, so only the source part enters code generation. Qian: move to the methodology}.
Then, an \textbf{Assembly Agent} builds the final model by aligning matched connector pairs using FreeCAD's constraint solver.
Because the connectors were already defined by the Design Agent, this step is strictly mathematical and does not require a VLM or LLM.
Finally, a \textbf{Review Agent} evaluates a generated CAD assembly by inspecting multi-view renders of each part and the keyframe sequences of the assembly's joint motions. 

We further improve the generation quality from two aspects. First, we introduce validation steps within both Generation and Assembly Agents, accompanied by a cross-stage rollback mechanism. Using error feedback derived from the validation step in each agent, this mechanism classifies the error as either a \texttt{DESIGN}-level or \texttt{CODE}-level failure. It then re-invokes only the responsible agent, thereby reducing redundant effort.
Second, a self-evolving experience store accumulates design knowledge derived from the Review Agent into a partitioned vector database after each task. This allows relevant knowledge to be retrieved and referenced in subsequent tasks.
%raising success rates on subsequent tasks through retrieval, without model fine-tuning.
Our contributions are summarized as follows:

%Our approach starts from the observation that assembly generation conflates two sub-problems with fundamentally different computational profiles: \emph{predicting how parts connect} and \emph{physically joining them}. Relationship prediction requires high-level functional and structural reasoning to determine which surfaces mate, the governing joint types, and the location of attachment points. Once these attachment points are explicitly defined as local coordinate frames, assembly execution reduces to a straightforward, deterministic rigid-body alignment. Because prediction is semantic and execution is purely mathematical, the key design choice is where in the pipeline to place this relationship prediction (\cref{fig:pipeline_comparison}).

\begin{itemize}[nosep,leftmargin=*]
%    \item \method{}, a training-free multi-agent system that generates articulated CAD assemblies with five typed joints from multimodal inputs. To our best knowledge, this is the first work to explore this task.
\item \textbf{A Multi-Agent Framework for Articulated CAD}: We propose \method{}, the first training-free multi-agent system to generate editable, articulated CAD assemblies from text and images. This fills the gap between single-part CAD generation and non-editable 3D reconstruction.
%\item A three-stage architecture---Design Agent, Part CodeGen Agents, and deterministic Assembly Agent---coordinated through a connector contract that decouples relationship prediction from geometry generation and assembly execution, with a derive mechanism for symmetric and repeated parts.
\item \textbf{Design-Stage Assembly Planning}: We propose predicting assembly relationships using the Connector during the design rather than assembly stage, bypassing the spatial reasoning limitations of current LLMs/VLMs.

%\item Multi-modal VLM validation, cross-stage rollback with \texttt{DESIGN}/\texttt{CODE} error classification, and a self-evolving partitioned experience store that accumulates design knowledge without model fine-tuning.
\item \textbf{Cross-Stage Feedback and Self-Evolution}: To improve generation quality, we introduce validation steps within the generation and assembly stages, accompanied by a cross-stage rollback mechanism. We also propose a self-evolving experience store that accumulates design knowledge for future tasks.
\item \textbf{New Benchmark and Downstream Applications}: We introduce ArtiCAD-Bench, a 120-task benchmark, and conduct cross-domain evaluations on CADPrompt~\cite{alrashedy2025cadcodeverify}  and ACD~\cite{iliash2024s2o}.
In practice, \method{} supports requirement-driven conceptual design and generates CAD assemblies that are directly exportable as URDFs for robotic simulation.
%\item Requirement-driven conceptual design from functional specifications, generating structurally diverse options; URDF export for robotic simulation; and ArtiCAD-Bench, a 120-task benchmark with cross-domain evaluation on CADPrompt~\cite{alrashedy2025cadcodeverify} and ACD~\cite{iliash2024s2o}.
\end{itemize}

%% file: sec/related_work.tex
% ===============================================================
% RELATED WORK
% ===============================================================
\section{Related Work}
\label{sec:related}

\subsection{Parametric CAD Generation}

Early methods model discrete sketch-and-extrude sequences, with generative approaches like DeepCAD~\cite{wu2021deepcad} (autoregressive modeling), SkexGen~\cite{xu2022skexgen} (disentangled codebooks), Text2CAD~\cite{khan2024text2cad} (prompt-conditioned), CAD-GPT~\cite{wang2025cadgpt} (3D positional tokens), and CAD-Llama~\cite{li2025cadllama} (fine-tuning). Reconstruction counterparts include TransCAD~\cite{dupont2024transcad} and CAD-SIGNet~\cite{khan2024cadsignet}. To overcome the geometric limitations of fixed vocabularies, recent methods synthesize executable programmatic scripts (\eg, CadQuery). CAD-Recode~\cite{rukhovich2025cad} and Text-to-CadQuery~\cite{xie2025textcadquery} generate scripts from point clouds and text. CAD-Coder~\cite{guan2025cadcoder} introduces geometric rewards, while CADCrafter~\cite{chen2025cadcrafter} leverages latent diffusion. Advanced alignments are achieved via reinforcement learning in Cadrille~\cite{kolodiazhnyi2025cadrille} or sequential fine-tuning in CADmium~\cite{govindarajan2026cadmium}. 

To mitigate execution errors, several frameworks wrap LLMs in generate--execute--repair loops, utilizing sandbox traces (CADCodeVerify~\cite{alrashedy2025cadcodeverify}), visual feedback (Seek-CAD~\cite{seekcad2025}), evolutionary search (EvoCAD~\cite{preintner2025evocad}), or VLM-guided edits (CADEvolve~\cite{elistratov2026cadevolve}). Crucially, all these methods are restricted to producing single, static parts without multi-part hierarchies or joints.

\subsection{CAD Datasets and Assembly}

Common CAD datasets provide single-part histories (DeepCAD~\cite{wu2021deepcad}, Fusion 360 Gallery~\cite{willis2021fusion}, WHUCAD~\cite{fan2025whucad}) or multimodal annotations (Omni-CAD~\cite{xu2024cadmllm}, CADInstruct~\cite{lv2025cadinstruct}). While assembly datasets exist (\eg., Fusion 360 and AutoMate~\cite{jones2021automate}), works utilizing them primarily focus on joint-axis prediction (JoinABLe~\cite{willis2022joinable}), mate-type classification (AutoMate), or next-part recommendation~\cite{liang2024customizing} for \emph{pre-existing} components. Even CADKnitter~\cite{le2025cadknitter}, which generates a complementary part under geometric constraints, assumes a given base model. To our knowledge, \method{} is the first framework to synthesize complete, multi-part articulated CAD assemblies with typed degrees of freedom from scratch using high-level specifications.

\subsection{Articulated Object Generation and Reconstruction}

A separate body of work targets 3D articulated objects. Datasets like PartNet~\cite{mo2019partnet} and PartNet-Mobility~\cite{xiang2020sapien} provide fine-grained segmentations and joint annotations. For generation, methods learn diffusion priors (NAP~\cite{lei2023nap}), use graph-conditioned diffusion (SINGAPO~\cite{liu2025singapo}), leverage VLM agents to articulate existing meshes (Articulate-Anything~\cite{le2025articulateanything}), or synthesize geometry and articulation via compact 3D/latent tokens (PhysX-Anything~\cite{cao2026physxanything}, PAct~\cite{liu2026pact}). For reconstruction, approaches build digital twins from interactions or multi-view images (Ditto~\cite{jiang2022ditto}, PARIS~\cite{liu2023paris}), recover objects via Gaussian splatting (ArtGS~\cite{liu2025artgs}, GaussianArt~\cite{shen2026gaussianart}), or convert static meshes to openable ones (S2O~\cite{iliash2024s2o}). 

Critically, these methods produce non-editable mesh or Gaussian representations and generalize poorly beyond their training categories (primarily furniture). Unlike prior works, \method{} generates parametric, editable CAD assemblies from scratch without category restrictions.

\subsection{LLM-Based Multi-Agent Systems}

Multi-agent frameworks split complex tasks among role-specialized agents, outperforming monolithic systems in software engineering (MetaGPT~\cite{hong2024metagpt}, AutoGen~\cite{wu2024autogen}, ChatDev~\cite{qian2024chatdev}) and reasoning (ReAct~\cite{yao2023react}). For code-centric tasks, agents benefit from self-debugging (Self-Debug~\cite{chen2024selfdebug}), executable actions (CodeAct~\cite{wang2024codeact}), and verbal self-reflection (Reflexion~\cite{shinn2023reflexion}). Furthermore, agent decisions are routinely grounded via retrieval-augmented generation (RAG~\cite{lewis2020retrieval}, Self-RAG~\cite{asai2024selfrag}) and automated multimodal evaluation (VLM-as-a-Judge~\cite{chen2024mllm}). 

\method{} builds on these foundations with three mechanisms tailored to CAD assembly: a \emph{connector contract} that decouples relationship prediction from geometry generation; a \texttt{DESIGN}/\texttt{CODE} cross-stage rollback that localizes spatial errors to the responsible agent; and a self-evolving experience store that accumulates design knowledge across tasks without model fine-tuning.

%% file: sec/representation.tex
% ===============================================================
% ARTICULATED ASSEMBLY REPRESENTATION
% ===============================================================
\section{Articulated Assembly Representation}
\label{sec:representation}

\noindent\textbf{CAD Backend and Geometric Representation.}
We implement assemblies in FreeCAD~\cite{freecad2024}, an open-source parametric CAD platform.
Existing code-based CAD generation methods mostly target CadQuery~\cite{cadquery,rukhovich2025cad,xie2025textcadquery,guan2025cadcoder,chen2025cadcrafter,kolodiazhnyi2025cadrille,alrashedy2025cadcodeverify} or OpenSCAD~\cite{openscad}; CadQuery offers static placement but no joint types or kinematic solving, and OpenSCAD is purely CSG-based with no assembly concept.
FreeCAD combines sketch-based feature modeling (pad, pocket, revolve, loft) with a constraint-based Assembly solver. 
Mathematically, a generated Python script $\mathcal{G}_i$ is evaluated by the underlying OpenCASCADE kernel into a Boundary Representation (B-rep) solid. This solid comprises a set of topological entities $\mathcal{T}_i$ (\eg., faces, edges, vertices). As will be detailed in \cref{subsec:connector}, directly referencing these entities across parts is  unstable; therefore, our representation abstracts connection interfaces into explicit local coordinate frames $\mathbf{c} \in SE(3)$ (representing 3D rigid transformations) instead of implicit topological faces.

\begin{figure}[t]
  \centering
  \includegraphics[width=\linewidth]{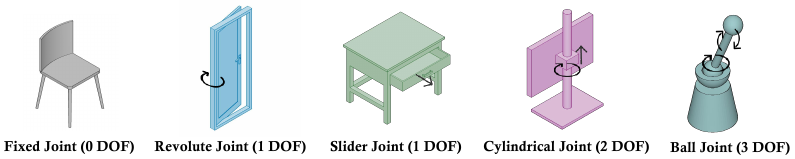}
  \vspace{-2em}
  \caption{The five core kinematic joint types utilized in \method{}. Each joint connects two parts at a shared coordinate frame; the specific degrees of freedom (DOF) constraints determine the allowed relative motion, which is subsequently resolved by FreeCAD's Assembly solver.}
  \label{fig:freecad}
  \vspace{-1.5em}
\end{figure}

\noindent\textbf{Kinematic Joint Formulation.}
Instead of defining a joint as a coincidental constraint between dynamic topological surfaces $t_i \in \mathcal{T}_i$ and $t_j \in \mathcal{T}_j$, \method{} defines a joint as a kinematic constraint between two explicit local frames $\mathbf{c}_i$ and $\mathbf{c}_j$. 
In our implementation, we utilize five core kinematic joint types (including \texttt{Fixed}, \texttt{Revolute}, \texttt{Slider}, \texttt{Cylindrical}, and \texttt{Ball}), all scriptable through its Python API in headless mode (\cref{fig:freecad}).
Given these typed constraints, the solver computes deterministic rigid body transformations in $SE(3)$ that satisfy them, completely decoupling assembly from the underlying B-rep topologies.

\noindent\textbf{From Assembly Graph to Kinematic Tree.}
An articulated assembly naturally forms a graph whose edges are joints, but graphs with closed loops impose coupled constraints that are difficult for an LLM to keep consistent.
We restrict the structure to a \emph{kinematic tree} $\mathcal{T} = (\mathcal{P}, \mathcal{J}, g)$---the standard articulation representation in robotics and embodied AI (\eg \ URDF).
$N$ parametric parts $\mathcal{P} = \{p_i\}_{i=1}^{N}$ are connected by $N{-}1$ typed joints $\mathcal{J} = \{j_k\}_{k=1}^{N-1}$, rooted at a ground part $g \in \mathcal{P}$ (fixed to the world frame).
Each joint carries a type $\tau_k$ from FreeCAD's five supported types and optional motion limits.
The total degrees of freedom are:
\vspace{-8pt}
\begin{equation}
    D = \sum_{k=1}^{N-1} \delta(\tau_k), \quad \delta(\tau) = \begin{cases} 0 & \tau = \texttt{Fixed} \\ 1 & \tau \in \{\texttt{Revolute}, \texttt{Slider}\} \\ 2 & \tau = \texttt{Cylindrical} \\ 3 & \tau = \texttt{Ball} \end{cases}
    \label{eq:dof}
\end{equation}
The acyclic topology ensures each part has a unique parent, so the solver always receives a well-posed problem.

%% file: sec/method.tex
% ===============================================================
% METHOD
% ===============================================================
\section{Method}
\label{sec:method}

\begin{figure}[t]
  \centering
  \includegraphics[width=\linewidth]{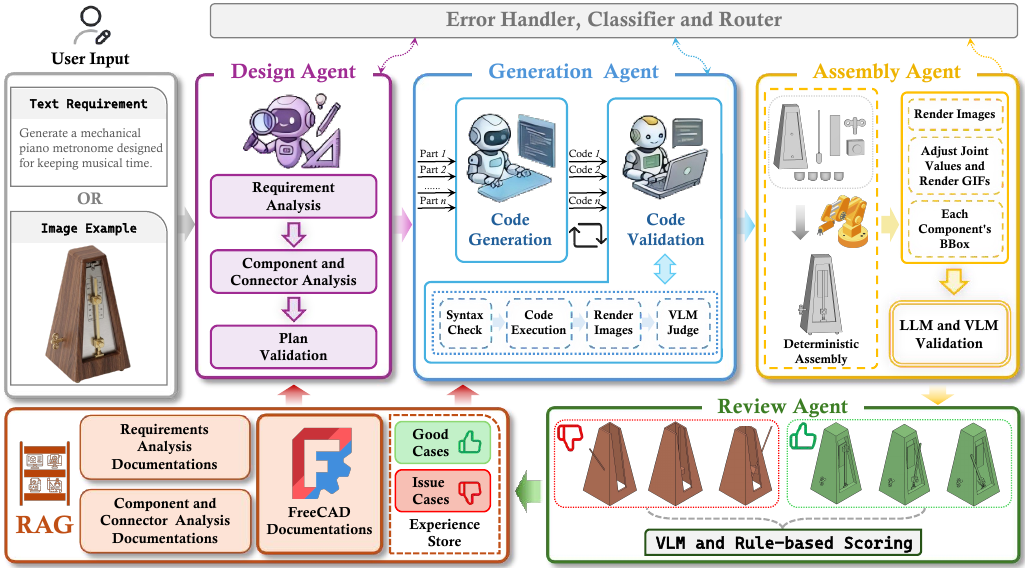}
  \vspace{-2em}
  \caption{Overview of the \method{} pipeline.
  A Design Agent decomposes multimodal input into components and connectors;
  Generation Agents generate per-part FreeCAD scripts through a generate--execute--repair loop with VLM validation;
  a deterministic Assembly Agent aligns parts and verifies the result via VLM and LLM judges;
  a Review Agent scores the output and records the case into the partitioned experience store.
  Cross-stage rollback (top) propagates failures to the responsible stage.}
  \label{fig:pipeline}
  \vspace{-1.5em}
\end{figure}

\textbf{Overview.}
As illustrated in \cref{fig:pipeline}, \method{} mirrors a design--production--assembly workflow.
A \emph{Design Agent} produces a structured plan with connector specifications; \emph{Generation Agents} produce each part independently; a deterministic \emph{Assembly Agent} joins them; and a \emph{Review Agent} scores the result and feeds review into an experience store.

\subsection{Connector Contract}
\label{subsec:connector}

The \emph{connector} is a named attachment point on a part, implemented as a local coordinate frame with a semantic label:
\begin{equation}
    c = (n,\; \mathbf{o} \in \mathbb{R}^3,\; \hat{\mathbf{z}} \in \mathbb{S}^2,\; \hat{\mathbf{x}} \in \mathbb{S}^2,\; l),
    \label{eq:connector}
\end{equation}
where $n \in \mathcal{N}$ is the unique identifier name, $\mathbf{o}$ is the origin in part-local coordinates, $\hat{\mathbf{z}}$ the primary axis (rotation or slide direction), $\hat{\mathbf{x}}$ an orthogonal reference, and $l \in \mathcal{L}$ describes the attachment's semantic purpose from a label space $\mathcal{L}$. Each joint in the kinematic tree (Sec. 3) references one connector on each of its two parts.
Connectors serve as a cross-stage contract: the Design Agent specifies them at plan time, Generation Agents realize the corresponding frames on constructed solids, and the Assembly Agent aligns matched pairs through FreeCAD's joint solver without any LLM call.
By fixing connectors early, assembly reduces to deterministic frame alignment, and a failed part can be regenerated in isolation.

\noindent\textbf{Justification.} 
To formalize the advantage of early prediction, consider an assembly graph of $N$ parts and $E$ joints. In late prediction, the LLM must match generated topological entities (\eg, B-rep faces) across parts, resulting in a combinatorial search space $\mathcal{O}(V^{2|E|})$, where $V$ is the average number of topological features per part. Furthermore, parametric CAD systems suffer from the \textit{Topological Naming Problem} (TNP)—minor code edits unpredictably re-index entities, making cross-part mapping highly volatile. By establishing a set of connector contracts $\mathcal{C} = \{\mathbf{c}_i\}_{i=1}^N$ \textit{a priori}, \method{} collapses this search space to $\mathcal{O}(1)$ deterministic frame alignment. 

Probabilistically, the contract acts as a Markov blanket. Rather than a monolithic joint distribution where a single error triggers cascading global failures (with expected retries scaling to $\mathcal{O}(K^N)$, where $K$ is the expected retries per part), the part generations become conditionally independent: 
\vspace{-4pt}
\begin{equation}
P(\mathcal{G}_1, \dots, \mathcal{G}_N \mid \mathcal{C}) = \prod_{i=1}^N P(\mathcal{G}_i \mid \mathbf{c}_i).
\end{equation}
This mathematical decoupling isolates failures, bounding the expected rollback cost linearly to $\mathcal{O}(N \cdot K)$.

\subsection{Design Agent}
\label{subsec:planning}

The Design Agent converts multimodal input into a structured plan.

\noindent\textbf{Requirement Analysis.}
A VLM-based module parses text and reference images into a specification of components, spatial relations, and constraints.
For under-specified inputs, a brainstorm module proposes structurally distinct alternatives for the user to choose from.

\noindent\textbf{Plan Generation.}
Given the specification and similar past plans from the experience store (\cref{subsec:rag}), the Design Agent outputs a declarative plan:
\begin{equation}
    \mathcal{P} = \bigl(\{p_i\}_{i=1}^{N},\; \{j_k\}_{k=1}^{N-1},\; g,\; D\bigr),
    \label{eq:plan}
\end{equation}
where each component $p_i$ carries design parameters, an orientation hint, and \emph{reference} connectors $\{c_{i,m}\}$; each joint $j_k$ pairs two connectors with a type and motion limits; and $D$ is the declared total DOF.
The plan is validated structurally: the joint graph must form a tree rooted at $g$, and $D$ must match \cref{eq:dof}.

\noindent\textbf{Derive Mechanism.}
Symmetric or repeated parts---\eg two refrigerator doors (mirror) or four table legs (rigid translation)---are handled by designating one as the source and specifying a deterministic $SE(3)$ transform for each copy, so only the source enters code generation.
\subsection{Generation Agents}
\label{subsec:codegen}

A Generation Agent is spawned for each non-derived component in the plan.

\noindent\textbf{Geometry Construction.}
The agent generates a FreeCAD Python script to model the part's geometry.
The script executes in a sandboxed FreeCAD process; on failure, the error trace and a render of the partial geometry feed back into the agent for repair via a generate--execute--repair loop.
Derived parts bypass the LLM entirely: a deterministic script simply applies the planned $SE(3)$ transform to the source geometry and its connectors.

\noindent\textbf{Connector Realization.}
Rather than forcing rigid coordinates blindly, the Connector Contract acts as a semantic spatial constraint. 
During code generation, the Generation Agent fine-tunes the exact placement of the exported connector frames based on the actual shape and dimensions of the locally generated geometry. 
This ensures each connector is accurately positioned at a semantically valid location---such as the exact center of a cylinder's top face---without breaking the global contract.

\noindent\textbf{Local Validation.}
After successful execution, a VLM compares multi-view renders of the generated part against the initial specification, checking shape, proportion, and orientation.
If a mismatch is found, the local VLM feedback is sent to the central error handler (detailed in \cref{subsec:rollback}) to determine the next steps.

\subsection{Assembly Agent}
\label{subsec:assembly}

The Assembly Agent synthesizes the complete kinematic tree using the realized components.

\noindent\textbf{Deterministic Assembly.}
Given the generated parts and their exported connector frames, a deterministic script computes rigid transforms to align each joint's connector pair. It then applies FreeCAD Assembly constraints to establish the joints. Because the topology was resolved by the Design Agent, no LLM or VLM is involved in this physical alignment step.

\noindent\textbf{Global Verification.}
The assembly is rigorously inspected via a VLM-LLM pipeline. A VLM first analyzes multi-view renders and motion keyframes for spatial and kinematic validity. An LLM judge then synthesizes these observations with bounding-box data and requirements to issue a structured verdict on placement, interference, and motion fidelity. Negative verdicts trigger the central error handler for resolution.

\subsection{Cross-Stage Rollback Mechanism}
\label{subsec:rollback}

The primary motivation for introducing VLM-based validation in the Generation (\cref{subsec:codegen}) and Assembly (\cref{subsec:assembly}) stages is to act as distributed sensors. To make these multi-stage feedback actionable without discarding successful intermediate output, \method{} employs a Cross-stage Rollback Mechanism (i.e., \emph{Error Handler, Classifier, and Router} in \cref{fig:pipeline}). 

\noindent\textbf{Error Classification and Routing.}
When a failure is reported by any VLM or LLM judge, the router analyzes the feedback to localize the defect. It classifies the failure as either a \texttt{CODE} error (the Python script failed to fulfill a valid specification, \eg, a hole is incorrectly sized) or a \texttt{DESIGN} error (the underlying connector plan is logically or physically flawed, \eg, parts heavily interfere after assembly). Based on this classification, the router invokes a \emph{cross-stage rollback}, routing the specific visual diagnostics and error traces back to either the responsible Generation Agent or the upstream Design Agent.

\noindent\textbf{Targeted Repair.}
%The true power of this rollback mechanism stems from the Connector Contract. 
With the Connector Contract, individual parts are generated conditionally independent (\cref{subsec:connector}) which enables the router to perform \emph{targeted repair}. It partitions the existing parts into \emph{keep}, \emph{regenerate}, and \emph{newly introduced} subsets. Faultless components are preserved in the ``keep'' pool, ensuring that only the affected nodes in the kinematic tree are re-planned or re-generated. This structured error routing breaks the cycle of cascading failures and minimizes redundant LLM queries.

\subsection{Review Agent and Memory System}
\label{subsec:rag}

\method{} mitigates historical errors and API hallucinations via an evolving memory system curated by a dedicated Review Agent.

\noindent\textbf{Review Agent.}
Post-assembly, the \emph{Review Agent} performs VLM- and rule-based evaluation of the output's geometric fidelity and kinematic health. It then distills the full generation trace---encompassing requirements, connector plans, code, and repair trajectories---into a structured case summary.

\noindent\textbf{Experience Store.}
Summaries are partitioned into \emph{Good} or \emph{Issue} cases in FAISS~\cite{johnson2019faiss}. This enables an \emph{asymmetric retrieval strategy}: Design Agents derive both \emph{positive design heuristics} and \emph{negative constraints} from the respective partitions, while Generation Agents strictly use \emph{Good Cases} as clean \emph{few-shot templates}. This cycle improves success rates without fine-tuning.

\noindent\textbf{Documentation Store.}
To bridge the semantic gap between user intent and CAD API nomenclature, we employ \emph{intent-driven retrieval}. An LLM predicts probable API signatures from geometry; these embeddings query chunked documentation to supply agents with precise syntax.

%% file: sec/experiments.tex
% ===============================================================
% EXPERIMENTS
% ===============================================================
\section{Experiments}
\label{sec:experiments}

\subsection{Experimental Setup}
\label{subsec:setup}

\noindent\textbf{Benchmarks.}
We evaluate \method{} on three benchmarks:

\emph{(1) ArtiCAD-Bench (Ours):} A proposed comprehensive benchmark comprising 120 assembly tasks in two subsets. The first includes 90 diverse real-world designs (50 articulated, 40 static) spanning furniture, toys, and appliances, driven by varying modalities (30\% text, 30\% image, 40\% both). The second consists of 30 industrial assemblies curated from the Fusion~360 dataset~\cite{willis2022joinable}, ranging from 2 to 6 parts, conditioned on assembly and per-part reference images. All tasks are evaluated using our VLM-based scoring protocol.

\emph{(2) CADPrompt}~\cite{alrashedy2025cadcodeverify}: A 200-item text-to-CAD dataset used to verify that our assembly-oriented pipeline does not compromise single-part generation quality compared to dedicated single-part methods.

\emph{(3) ACD}~\cite{iliash2024s2o}: The Articulated Containers Dataset (354 objects). We compare \method{} against state-of-the-art single-image articulated reconstruction methods, focusing strictly on joint estimation and resting-state geometry metrics.

\noindent\textbf{Backbone Model.}
The default backbone for \method{} is Gemini-3-Flash~\cite{deepmind_gemini3flash_modelcard_2025}. 
%used for both LLM and VLM calls. 
We also report results using Gemini-3-Pro on ArtiCAD-Bench to show the effect of a stronger backbone.
On CADPrompt (\cref{subsec:cad_comparison}), both \method{} and Single-VLM Loop use Gemini-3-Pro to control for backbone capacity. 

\noindent\textbf{Single-VLM Loop Baseline.}
To isolate the contributions of our multi-agent architecture, we compare against a \emph{Single-VLM Loop} baseline: a single backbone VLM receives the full task description and generates a single FreeCAD Python script to create and assemble all parts in a single pass. 
%(\textit{i.e.}, without the connector contract, staged pipeline, rollback, or experience store).
This baseline uses the same generate-execute-repair loop (up to 5 retries) but has no design/code/assembly decomposition.
We evaluate this baseline using four backbones: GPT-5.2~\cite{openai_gpt52_systemcard_2025}, Claude-Opus-4.6~\cite{anthropic_claude_opus46_systemcard_2026}, Gemini-3-Flash~\cite{deepmind_gemini3flash_modelcard_2025}, and Gemini-3-Pro~\cite{deepmind_gemini3pro_modelcard_2025}.

\noindent\textbf{Evaluation Protocol.}
Because ArtiCAD-Bench tasks are open-ended designs without ground-truth CAD models, geometric metrics like Chamfer Distance are not applicable.
Instead, we adopt a VLM-based scoring protocol inspired by G-Eval~\cite{liu2023geval}, MLLM-as-a-Judge~\cite{chen2024mllm}, and CAD-Judge~\cite{zhou2025cadjudge}.
Each generated assembly is rendered from multiple viewpoints (including joint motion keyframes for articulated models) and evaluated independently by three frontier VLMs: GPT-5.2~\cite{openai_gpt52_systemcard_2025}, Claude-Opus-4.6~\cite{anthropic_claude_opus46_systemcard_2026}, and Gemini-3-Pro~\cite{deepmind_gemini3pro_modelcard_2025}.

Each judge follows a chain-of-thought process~\cite{liu2023geval}: (1)~describe the observed parts and spatial layout from the renders, (2)~compare geometry and detail against the specification and reference images, (3)~analyze joint motion from keyframe sequences if present, and (4)~assign scores based on the rubric below.
To minimize subjective drift, the judges output structured JSON files containing their per-dimension reasoning and final integer scores.

We adopt three 1--5 Likert metrics:
\emph{Geometry} (Geo.) checks shape accuracy;
\emph{Detail} assesses feature coverage;
and \emph{Motion} evaluates kinematic correctness (defaulting to 5 for static items).
\emph{Success} (Succ.) is binary (0/1), marking valid compilation with all parts.
Final scores average three judges~\cite{zheng2023judging}.
Reliability is robust: Krippendorff's $\alpha$ ranges from 0.58 to 0.64, and the 3-rater mean ICC$(2,3)$ exceeds 0.81 across dimensions, confirming the stability of our VLM-based evaluation.
Standard geometric metrics are used for CADPrompt and ACD benchmarks.

\subsection{Main Results and Ablations on ArtiCAD-Bench}
\label{subsec:main_results}

\begin{table}[t]
\centering
\caption{Main results on ArtiCAD-Bench (120 items). Metrics defined in \cref{subsec:setup}; scores averaged across three VLM judges. For static assemblies, the Motion metric is not included in the average calculation. Best in \textbf{bold}, second best \underline{underlined}}
  \vspace{-1em}
\label{tab:main_results}
\setlength{\tabcolsep}{5pt}
\resizebox{0.8\linewidth}{!}{
\begin{tabular}{l cccc}
\toprule
Method & Geo.\ $\uparrow$ & Detail $\uparrow$ & Motion $\uparrow$ & Succ.\ $\uparrow$ \\
\midrule
Single-VLM Loop (Claude-Opus-4.6)      & 3.13 & 2.71 & 3.66 & \textbf{100\%} \\
Single-VLM Loop (GPT-5.2)              & 3.14 & 2.87 & 3.50 & 98.3\% \\
Single-VLM Loop (Gemini-3-Flash)         & 3.06 & 2.58 & 3.53 & \underline{99.2\%} \\
Single-VLM Loop (Gemini-3-Pro)         & 3.31 & 2.82 & 3.67 & 98.3\% \\
\method{} (Gemini-3-Flash)             & \underline{3.41} & \underline{2.92} & \underline{3.82} & \textbf{100\%} \\
\method{} (Gemini-3-Pro)               & \textbf{3.57} & \textbf{3.14} & \textbf{3.91} & \textbf{100\%} \\
\bottomrule
\end{tabular}}
  \vspace{-1.5em}
\end{table}

\noindent\textbf{Ablation Studies.}
We ablate key components of \method{} on ArtiCAD-Bench to measure their individual contributions:

\begin{itemize}[nosep]
     \item[(a)] \textbf{Late prediction of assembly relationship}: In this variant, the Design Agent specifies only the part list and descriptions, without predicting assembly relationship (\textit{i.e.}, connectors). The Generation Agents produce geometry code exclusively for individual parts. Connection planning is deferred to the Assembly Agent, where an LLM interprets the generated parts, produces assembly constraints, and completes the assembling process. Validation and the experience store remain the same as \method{}.
    \item[(b)] \textbf{w/o cross-stage rollback}:
    This variant removes the VLM-based validation from both the Generation and Assembly Agents, along with the cross-stage rollback mechanism it triggers. Errors are caught only through execution failures rather than visual inspection. This isolates and verifies the effectiveness of VLM-based visual validation combined with cross-stage rollback. 
    %This variant removes VLM-based validation of both individual parts and the final assembly, along with the cross-stage rollback it triggers. Errors are caught only by execution failures, not by visual inspection.
    \item[(c)] \textbf{w/o experience store}: Compared to \method{}, this variant removes the experience store from retrieval augmentation process while keeping the documentation store, isolating the contribution of accumulated design knowledge gained from prior tasks.
\end{itemize}

\begin{table}[t]
\centering
\caption{Ablation study on ArtiCAD-Bench. All ablation variants use Gemini-3-Flash as the backbone. Each variant removes one key component. Metrics follow \cref{tab:main_results}. Avg.\ Iter.\ counts the mean generate-execute-repair iterations per task (lower is better).}
  \vspace{-1em}
\label{tab:ablation}
\resizebox{\linewidth}{!}{
\setlength{\tabcolsep}{4pt}
\begin{tabular}{l cccc c}
\toprule
Variant & Geo.\ $\uparrow$ & Detail $\uparrow$ & Motion $\uparrow$ & Succ.\ $\uparrow$ & Avg.\ Iter.\ $\downarrow$ \\
\midrule
\method{}                               & 3.41& 2.92& 3.82&  100\% & 3.1 \\
(a) Late prediction of assembly relationship  & 3.11& 2.65& 3.16& 89.2\% & --  \\
(b) w/o cross-stage rollback              & 3.15& 2.89& 3.63& 95.0\% & --  \\
(c) w/o experience store              & 3.37& 2.94& 3.77&  100\% & 4.4 \\
\bottomrule
\end{tabular}
}
  \vspace{-1em}
\end{table}

\subsection{Comparison with CAD Code Generation Methods}
\label{subsec:cad_comparison}

We evaluate on CADPrompt~\cite{alrashedy2025cadcodeverify} to verify that the assembly-oriented design does not degrade its performance on single-part tasks.
Both \method{} and Single-VLM Loop use Gemini-3-Pro here, following the Refine-2 protocol of CADCodeVerify~\cite{alrashedy2025cadcodeverify} with two refinement iterations.
The model weights and inference code of 3D-PreMise~\cite{yuan2024premise}, CADCodeVerify~\cite{alrashedy2025cadcodeverify}, and Seek-CAD~\cite{seekcad2025} are not publicly available; we report their published numbers.

\noindent\textbf{Metrics.}
We sample 1{,}000 points from each mesh, apply Iterative Closest Point (ICP) alignment, and normalize into the unit cube.
We report three metrics: \emph{Point Cloud Distance} (PCD, symmetric Chamfer distance), \emph{Hausdorff Distance} (HD), and \emph{Intersection-over-Ground-Truth} (IoGT, bounding-box volume overlap).
Failed samples receive worst-case scores ($\mathrm{PCD}=\mathrm{HD}=\sqrt{3}$, \ie,the unit-cube diagonal; $\mathrm{IoGT}=0$).

\begin{table}[t]
\centering
\caption{Comparison on CADPrompt. $^\dagger$Results from the respective papers (model weights and inference code are not publicly available). Best in \textbf{bold}.}
  \vspace{-1em}
\label{tab:cadprompt}
\resizebox{0.8\linewidth}{!}{
\setlength{\tabcolsep}{3.5pt}
\begin{tabular}{l cc cc cc c}
\toprule
\multirow{2}{*}{Method}
& \multicolumn{2}{c}{IoGT $\uparrow$} & \multicolumn{2}{c}{PCD $\downarrow$} & \multicolumn{2}{c}{HD $\downarrow$} & \multirow{2}{*}{Compile Rate $\uparrow$} \\
\cmidrule(lr){2-3} \cmidrule(lr){4-5} \cmidrule(lr){6-7}
& mean & med. & mean & med. & mean & med.  \\
\midrule
3D-PreMise$^\dagger$~\cite{yuan2024premise}             & -- & 0.942 & -- & 0.137 & -- & 0.446 & 91.0\% \\
CADCodeVerify$^\dagger$~\cite{alrashedy2025cadcodeverify} & -- & 0.944 & -- & 0.127 & -- &  0.419 & 96.5\% \\
Seek-CAD$^\dagger$~\cite{seekcad2025}                   & 0.801 & -- & 0.199 & -- & 0.538 & -- & -- \\
\midrule
Single-VLM Loop         & 0.873 & 0.979 & 0.044 & 0.027 & 0.148 & 0.103 & 99.5\% \\
\method{} (ours)         & \textbf{0.897} & \textbf{0.986} & \textbf{0.034} & \textbf{0.025} & \textbf{0.130} & \textbf{0.090} & \textbf{100.0\%} \\
\bottomrule
\end{tabular}
}
  \vspace{-1em}
\end{table}

The controlled comparison with Single-VLM Loop (same backbone, no multi-agent pipeline) shows that the structured planning in \method{} does not hurt, even slightly improves single-part quality.

% TODO: Qualitative comparison figure: rendered parts from ArtiCAD vs.\ baselines.

\subsection{Comparison with Articulated Object Methods}
\label{subsec:articulated_comparison}

We compare \method{} against three representative articulated object methods on the ACD dataset~\cite{iliash2024s2o}: \textbf{first}, \textbf{SINGAPO}~\cite{liu2025singapo} predicts part attributes and kinematics from a single image via diffusion, subsequently assembling the object through mesh retrieval; \textbf{second}, \textbf{Articulate-Anything}~\cite{le2025articulateanything} employs a VLM to iteratively code articulation for retrieved meshes (evaluated under its single-image setting); \textbf{third}, \textbf{PAct}~\cite{liu2026pact} uses part-centric latent tokens to simultaneously synthesize geometry and motion feed-forwardly from a single image.

Prior to evaluation, we normalize the meshes by their bounding box diagonals and align them using Iterative Closest Point (ICP). We then report the following metrics: resting-state Chamfer distance (RS-CD, mean/median, $\downarrow$), resting-state IoU (RS-IoU, mean/median, $\uparrow$), movable joint type accuracy (Movable Type Acc., mean, $\uparrow$), and movable joint F1 (Movable F1, mean, $\uparrow$).

\begin{figure}[!t]
  \centering
  \includegraphics[width=\linewidth]{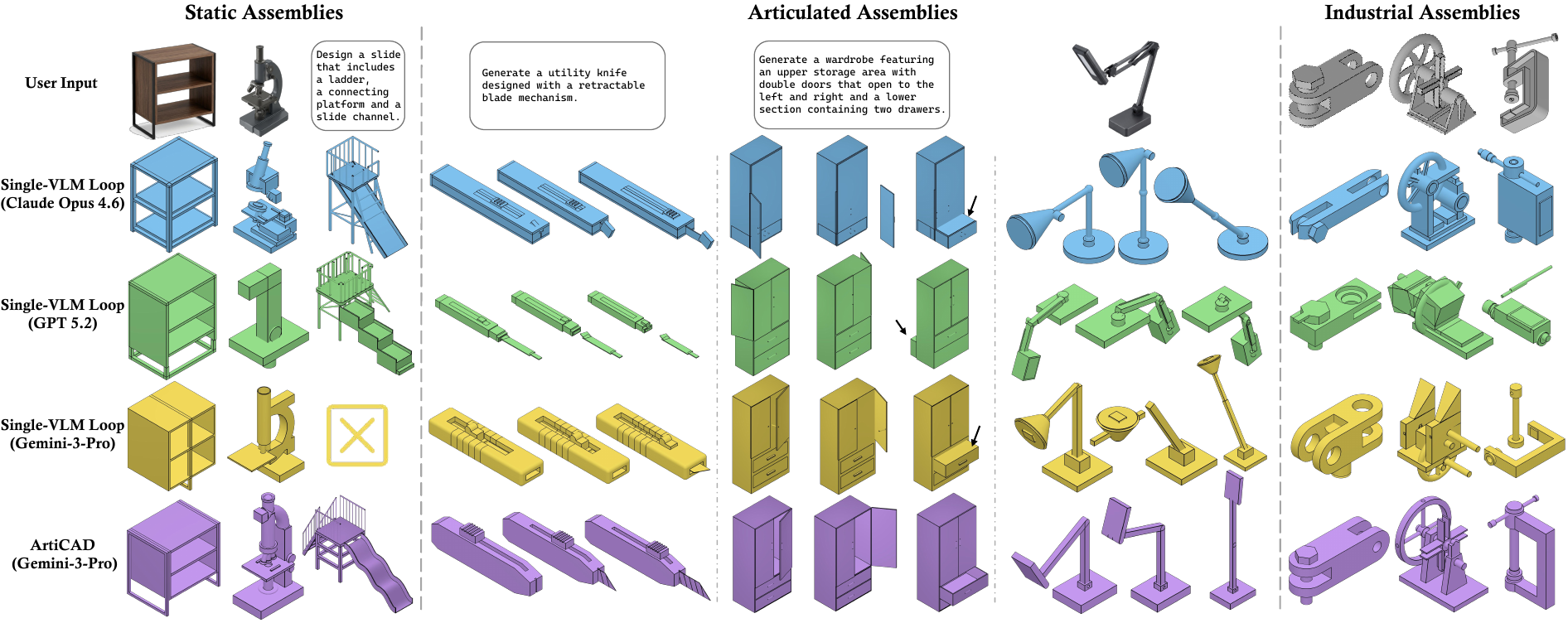}
  \vspace{-2em}
  \caption{Qualitative results comparing \method{} with Single-VLM Loop on our bench.}
  \label{fig:qualitative}
  \vspace{-0.5em}
\end{figure}

\begin{table}[t]
\centering
\caption{Comparison on ACD dataset. $^\dagger$Articulate-Anything uses Claude-Opus-4.6 as backbone; our method uses Gemini-3-Flash. Best in \textbf{bold}.}
  \vspace{-1em}
\label{tab:acd_comparison}
\resizebox{0.85\linewidth}{!}{
\setlength{\tabcolsep}{3.5pt}
\begin{tabular}{lcccccc}
\toprule
\multirow{2}{*}{Method} &
\multicolumn{2}{c}{RS-CD$\downarrow$} &
\multicolumn{2}{c}{RS-IoU$\uparrow$} &
\multirow{2}{*}{Mov. Type Acc$\uparrow$} &
\multirow{2}{*}{Mov. F1$\uparrow$} \\
\cmidrule(lr){2-3} \cmidrule(lr){4-5}
& mean & med & mean & med & & \\
\midrule
SINGAPO~\cite{liu2025singapo} &
0.037 & 0.030 & 0.156 & 0.136 & 0.772 & 0.590 \\
Articulate-Anything$^\dagger$~\cite{le2025articulateanything} &
0.087 & 0.078& 0.194 & 0.181 & 0.812& 0.577\\
PAct~\cite{liu2026pact} &
0.036 & 0.025 & 0.346 & 0.371 & 0.732 & 0.450 \\
\midrule
\method{} (ours) &
\textbf{0.030} & \textbf{0.017} & \textbf{0.386} & \textbf{0.406} & \textbf{0.934} & \textbf{0.841} \\
\bottomrule
\end{tabular}}
  \vspace{-1.5em}
\end{table}

\subsection{Qualitative Analysis}
\label{subsec:qualitative}

As shown in \cref{fig:qualitative}, as generation tasks become more complex, the Single-VLM Loop baseline tends to oversimplify part shapes and lose functional geometric details. In contrast, \method{} preserves structure-aware geometry and produces more complete, better-organized assemblies. 
Specifically, inter-part alignment is cleaner, connections and overall layouts are more consistent with the intended functionality, and cross-part inconsistencies are minimized. 
For example, as indicated by the black arrows in \cref{fig:qualitative}, our method correctly models the drawer as a hollowed-out component, whereas the baseline often collapses it into a solid block, ignoring expected manufacturing structures. 
Furthermore, for articulated objects, our results exhibit more coherent and intuitive motions, highlighting the advantage of \method{} in both geometric consistency and kinematic plausibility. 

\begin{figure}[t]
  \centering
  \includegraphics[width=\linewidth]{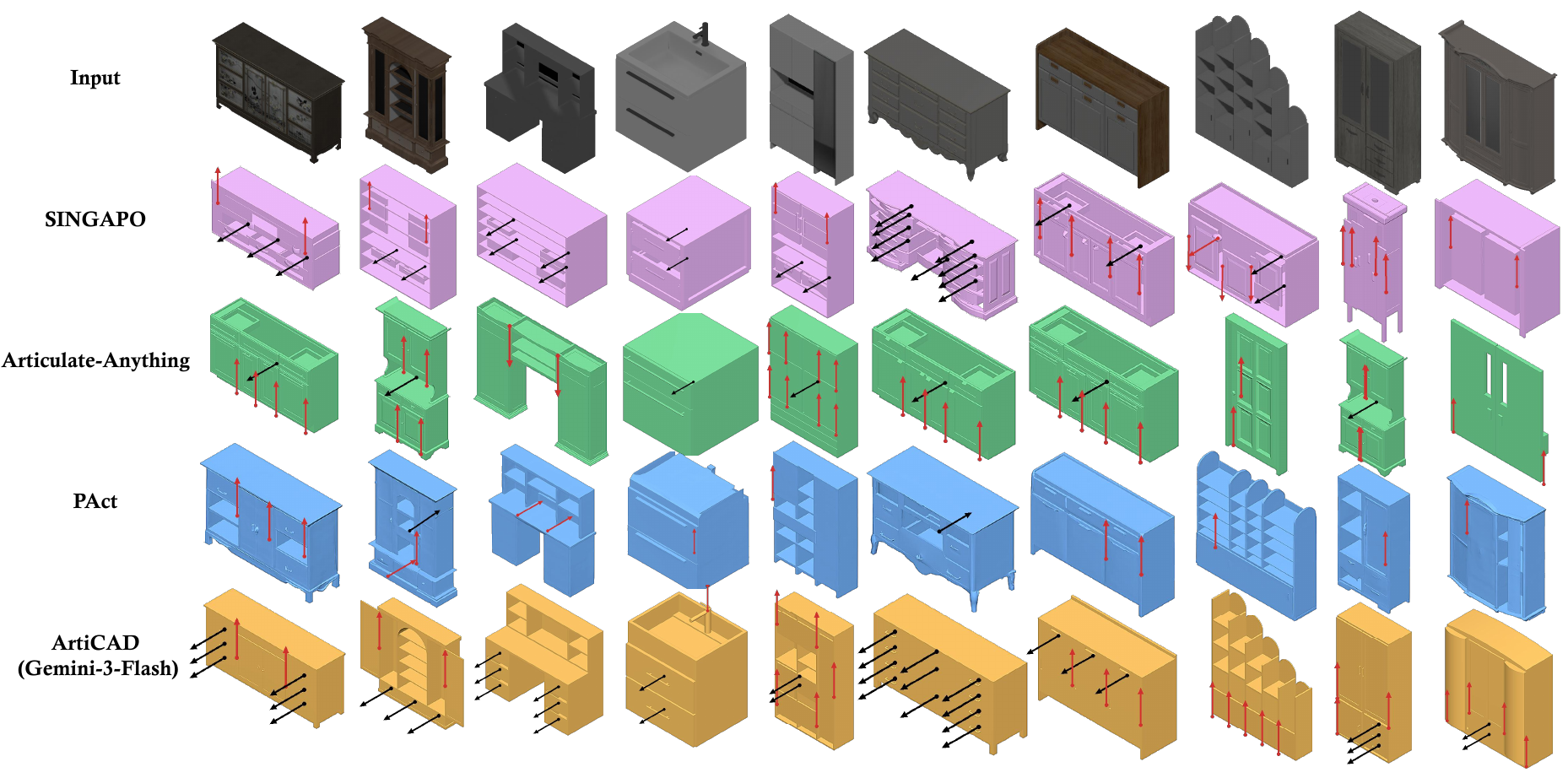}
  \caption{Qualitative comparisons between \method{} and SINGAPO, Articulate-Anything, and PAct on the ACD dataset. Black arrows indicate prismatic (translational) joints, and red arrows indicate revolute (rotational) joints.}
  \label{fig:mq2}
\end{figure}

On the ACD dataset, qualitative comparisons in \cref{fig:mq2} further reveal the distinct limitations of prior articulated object methods. Articulate-Anything, being retrieval-based, struggles with fine-grained structural variations and may retrieve nearly identical geometries for objects with similar global appearance but different door or drawer layouts. PAct often reconstructs geometric details well, but its joint prediction remains overly conservative and tends to miss valid movable joints. SINGAPO predicts joint types more reliably, yet frequently exhibits noticeable errors in joint localization. In contrast, \method{} better preserves the overall object shape while recovering a more complete set of joint types with more accurate spatial placement, leading to more coherent articulated structures and motions.

%% file: sec/applications.tex
% ===============================================================
% APPLICATIONS
% ===============================================================
\section{Applications}
\label{sec:applications}

Since \method{} generates parametric assemblies with typed joints and motion limits, its outputs serve use cases beyond static 3D content.

\noindent\textbf{Requirement-driven Design and Physical Prototyping.}
As illustrated in Fig.~\ref{fig:teaser}(Bottom), \method{} seamlessly bridges high-level conceptual design and physical prototyping. Given a functional prompt (\eg, ``Generate a tabletop double-person toy''), the brainstorm module proposes distinct structural candidates. For the selected \textit{Tabletop Football}, \method{} generates a fully articulated, fabrication-ready CAD assembly. The accompanying photos validate this pipeline, demonstrating the successful 3D printing and physical construction of the functional prototype.

%\noindent\textbf{Articulated Assets for Embodied AI.}
%Our pipeline automatically exports each assembled model as a URDF file with joint types and motion limits, ready for robotic simulation in environments such as SAPIEN~\cite{xiang2020sapien} or Isaac Sim. While existing articulated datasets have limited category coverage, \method{} generates novel, out-of-distribution object types on demand. Additional results are provided in the Supplementary.

\begin{figure}[t]
  \centering
  \includegraphics[width=\linewidth]{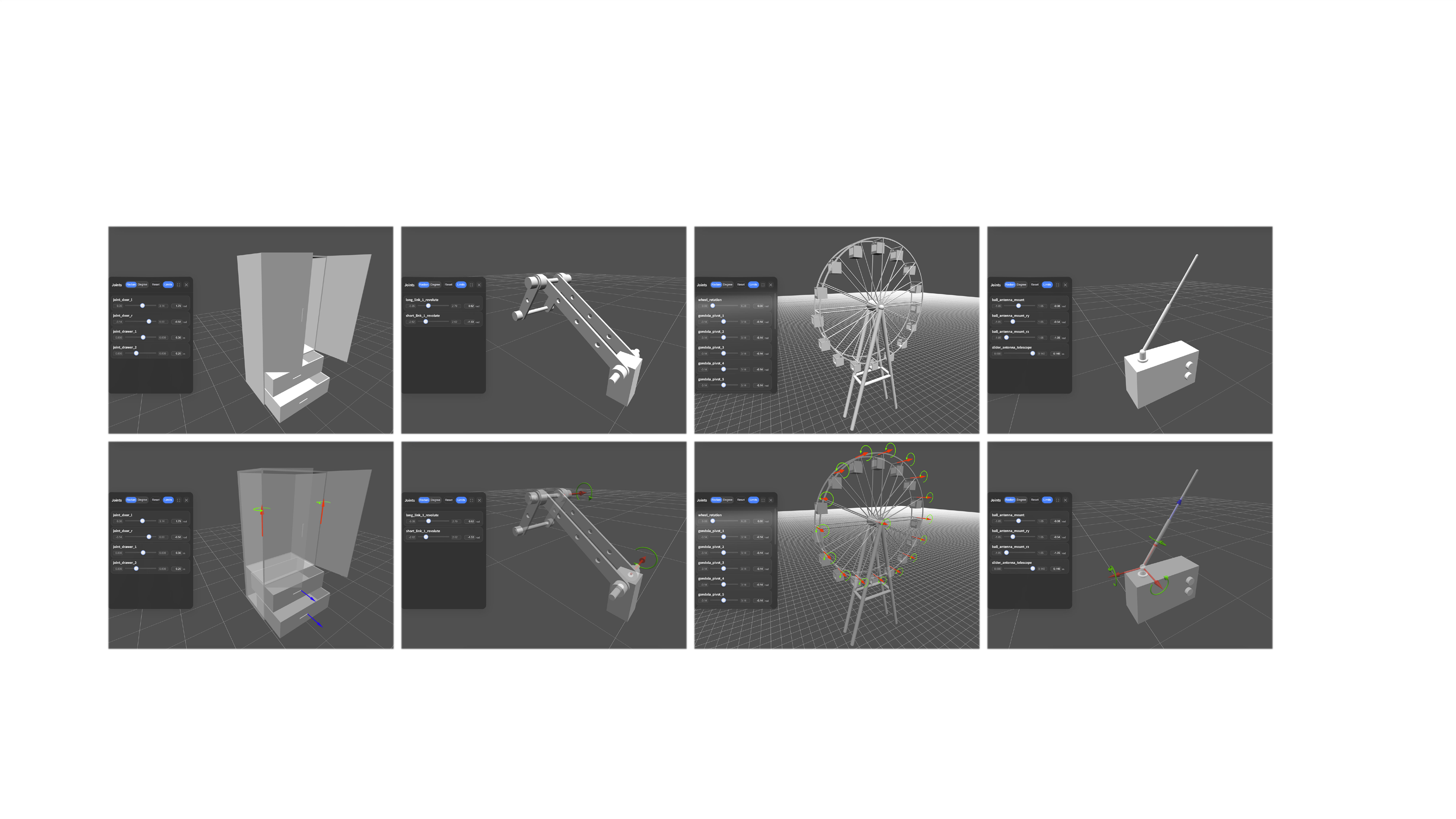}
  \vspace{-1.2em}
  \caption{URDF export verification for embodied AI applications. Top: exported
  assemblies loaded in Robot Viewer. Bottom: the same models with joint coordinate
  frames visualized. The exported URDFs preserve the intended joint structure, axis
  directions, and motion limits.}
  \label{fig:urdf_main}
  \vspace{-1.2em}
\end{figure}

\noindent\textbf{Articulated Assets for Embodied AI.}
Our pipeline automatically exports each assembled model as a URDF file with joint
types and motion limits, ready for robotic simulation in environments such as
SAPIEN~\cite{xiang2020sapien} or Isaac Sim~\cite{nvidia_isaac_sim}. As shown in
Fig.~\ref{fig:urdf_main}, visualization in Robot Viewer~\cite{fan2024robotviewer}
confirms that the exported joint structure, axis directions, and motion limits are
faithfully preserved. While existing articulated datasets have limited category
coverage, \method{} generates novel, out-of-distribution object types on demand.

% TODO: Figure — URDF in simulation environment

%\noindent\textbf{Kinematic skeleton for mesh refinement.}
%Beyond direct use, the generated assembly has potential as a kinematic skeleton: \method{} produces correct overall shape and joint kinematics but leaves surface detail coarse, as fillets, textures, and freeform curvature are limited by code-based geometry construction.
%Each part's mesh can be refined independently by downstream 3D generation or reconstruction methods, while the joint structure, motion limits, and contact surfaces at attachment points are kept as hard constraints.
%This would yield a high-fidelity model that remains fully articulated, combining the geometric precision of mesh-based methods with the kinematic correctness of \method{}.
%We leave full integration to future work, but the generated assemblies already provide the kinematic scaffolding that such pipelines require.

%% file: sec/conclusion.tex
% ===============================================================
% CONCLUSION
% ===============================================================
\section{Conclusion}
\label{sec:conclusion}

We presented \method{}, the first training-free multi-agent system that generates articulated CAD assemblies from multimodal inputs. 
By leveraging the \emph{connector contract}, \method{} decouples relationship prediction from geometry generation, simplifying the assembly process into a deterministic $\mathcal{O}(1)$ frame alignment. Furthermore, we enhanced the system's reliability and efficiency through a cross-stage rollback mechanism that precisely isolates design- and code-level errors, alongside a self-evolving experience store that accumulates knowledge for continuous improvement.  \method{} outperforms baselines across multiple benchmarks, yielding editable, simulation-ready CAD models.

\noindent\textbf{Limitations.}
First, the kinematic tree formulation cannot represent closed kinematic chains forming physical loops (\eg, scissor linkages or four-bar linkages). However, this acyclic constraint deliberately trades closed-loop complexity for deterministic, zero-hallucination assembly, successfully covering most everyday products. Second, like other multi-agent systems, our performance is fundamentally bounded by the general reasoning and code generation capabilities of underlying foundation models. Future work includes model fine-tuning and reinforcement learning on synthetic assembly trajectories.